\documentclass{article}

\usepackage{PRIMEarxiv}

\usepackage[utf8]{inputenc} %
\usepackage[T1]{fontenc}    %
\usepackage[colorlinks=true,linkcolor=blue,allcolors=blue]{hyperref}       %
\usepackage{url}            %
\usepackage{booktabs}       %
\usepackage{amsfonts}       %
\usepackage{nicefrac}       %
\usepackage{microtype}      %
\usepackage{fancyhdr}       %
\usepackage{graphicx}       %

\usepackage{wrapfig}

\usepackage{natbib}
\setcitestyle{authoryear,round,citesep={;},aysep={,},yysep={;}}

\usepackage{amsmath,amsfonts,bm}

\usepackage{graphicx}
\usepackage{algorithm}
\usepackage{algpseudocode}

\algrenewcommand\algorithmicrequire{\textbf{Hyperparameters:}}

\usepackage{tabularray}
\usepackage{tablefootnote}
\usepackage{enumitem}
\usepackage{multirow}

\newcommand{\doublecell}[2][c]{%
  \begin{tabular}[#1]{@{}c@{}}#2\end{tabular}}

\newcommand{\dd}[2]{\frac{\partial #1} {\partial #2}}
\newcommand{\dds}[2]{\frac{\partial^2 #1}{\partial {#2}^2}}
\newcommand{\der}[2]{\frac{d #1} {d #2}}

\newcommand{\Lpde}{\mathcal{L}_\text{PDE}}
\newcommand{\Lsup}{\mathcal{L}_\text{S}}
\newcommand{\Lb}{\mathcal{L}_\text{B}}
\newcommand{\Kpde}{K_\text{PDE}}
\newcommand{\Ks}{K_\text{S}}
\newcommand{\Kb}{K_\text{B}}

\def\vg{{\bm{g}}}

\def\vu{{\bm{u}}}
\def\U{{u}}

\newsavebox\CBox
\def\textBF#1{\sbox\CBox{#1}\resizebox{\wd\CBox}{\ht\CBox}{\textbf{#1}}}
\title{Improved Training of Physics-Informed\\ Neural Networks with Model Ensembles
}

\author{
  Katsiaryna Haitsiukevich \& Alexander Ilin \\
  \textit{Department of Computer Science} \\
  \textit{Aalto University} \\
  Espoo, Finland \\
  \texttt{\{firstname.lastname\}@aalto.fi} \\
}

\begin{document}
\maketitle

\begin{abstract}
Learning the solution of partial differential equations (PDEs) with a neural network is an attractive alternative to traditional solvers due to its elegance, greater flexibility and the ease of incorporating observed data. However, training such physics-informed neural networks (PINNs) is notoriously difficult in practice since PINNs often converge to wrong solutions.
In this paper, we address this problem by training an ensemble of PINNs.
Our approach is motivated by the observation that individual PINN models find similar solutions in the vicinity of points with targets (e.g., observed data or initial conditions) while their solutions may substantially differ farther away from such points. Therefore, we propose to use the ensemble agreement as the criterion for gradual expansion of the solution interval, that is including new points for computing the loss derived from differential equations. Due to the flexibility of the domain expansion, our algorithm can easily incorporate measurements in arbitrary locations. In contrast to the existing PINN algorithms with time-adaptive strategies, the proposed algorithm does not need a pre-defined schedule of interval expansion and it treats time and space equally. We experimentally show that the proposed algorithm can stabilize PINN training and yield performance competitive to the recent variants of PINNs trained with time adaptation.

\end{abstract}


\section{Introduction}

Partial differential equations (PDEs) are a powerful tool for modeling many real-world phenomena \citep{evans1998partial,courant89methods}. When derived from the first principles, partial differential equations can serve as predictive models which do not require any data for tuning. When learned from data, they often outperform other models by incorporating the inductive bias of the continuity of the modeled domain (time or space) \citep{chen2018neuralode,rubanova2019latent,iakovlev2021learning}. Inference in this type of models is done by solving partial differential equations, that is by finding a trajectory that satisfies the model equations and a set of initial and boundary conditions. Since analytic solutions exist only for a limited number of models (most likely derived from the first principles), inference is typically done by numerical solvers of differential equations.

One way of solving differential equations is to approximate the solution
by a neural network which is trained to satisfy a given set of differential equations, initial and boundary conditions. This approach is known in the literature under the name of \emph{physics-informed neural networks} \citep[PINNs, ][]{lagaris1998artificial,raissi2019pinn} and it can be seen as a machine learning alternative to %
classical numerical solvers. Despite the conceptual simplicity and elegance of the method, training PINNs is notoriously difficult in practice \citep{wang2021understanding,krishnapriyan2021characterizing}. It requires balancing of multiple terms in the loss function \citep{wang2021understanding,wang2021ntk} and the commonly used neural network architectures and parameter initialization schemes may not work best for PINNs~\citep{wang2021eigenvector,sitzmann2019siren}. 

\begin{figure*}[t]
\newcommand{\myheight}{36mm}
\centering
\begin{minipage}{0.31\linewidth}
\centering
\small Early stage
\\[1mm]
\includegraphics[height=\myheight,trim={16mm 6mm 43mm 10mm},clip]{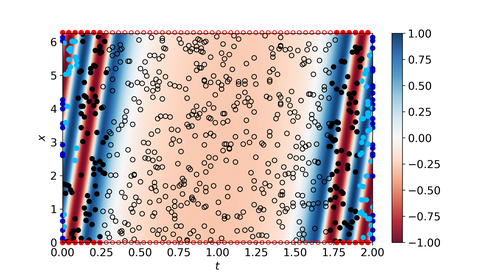}
\end{minipage}
\begin{minipage}{0.31\linewidth}
\centering
\small Intermediate results
\\[1mm]
\includegraphics[height=\myheight,trim={16mm 7mm 40mm 10mm},clip]{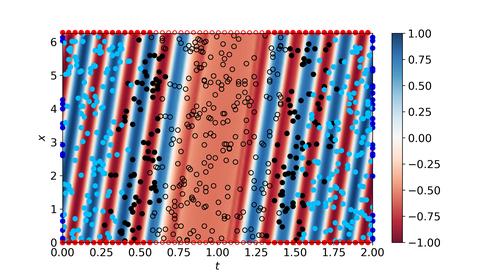}
\end{minipage}
\begin{minipage}{0.34\linewidth}
\centering
\small Final results
\\[1mm]
\includegraphics[height=\myheight,trim={16mm 7mm 21mm 10mm},clip]{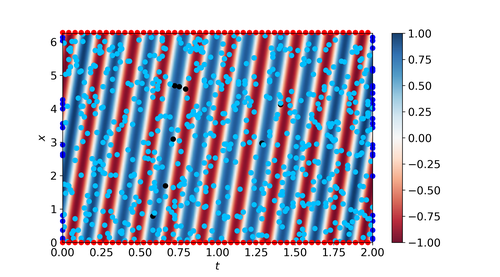}
\end{minipage}

\caption{The training procedure for the convection equation (Eqs.~\ref{eq:conv}--~\ref{eq:ibconv}) when the solution is known for $t = 0$ and $t = 2$. Gradual expansion of the solution domain (represented by the blue and black points) from both ends of the interval is possible due to the flexibility of the proposed approach.
The dot color scheme is from Fig.~\ref{f:alg}.
}
\label{f:conv_02_ens}
\end{figure*}

It has been noted by many practitioners that training PINNs often results in convergence to bad solutions \citep[see, e.g.,][]{krishnapriyan2021characterizing,sitzmann2019siren,wang2022respecting}.
Several recent works \citep{wight2020solving,krishnapriyan2021characterizing,mattey2022novel} address this problem by splitting the time interval into sub-intervals and sequentially training PINNs on each sub-interval.
This idea is often referred in the literature as time adaptation or time marching. The time-adaptive strategies make the training procedure of PINNs similar to classical numerical solvers which compute the solution gradually moving from the initial conditions towards the other end of the time interval. Similarly to classical solvers, many existing PINN algorithms are based on a pre-defined schedule of time adaptation.

\begin{wrapfigure}{r}{0.6\textwidth}
  \begin{center}
    \vskip -5mm
    \includegraphics[width=0.6\textwidth,trim={3mm 2mm 6.5mm 6mm},clip]{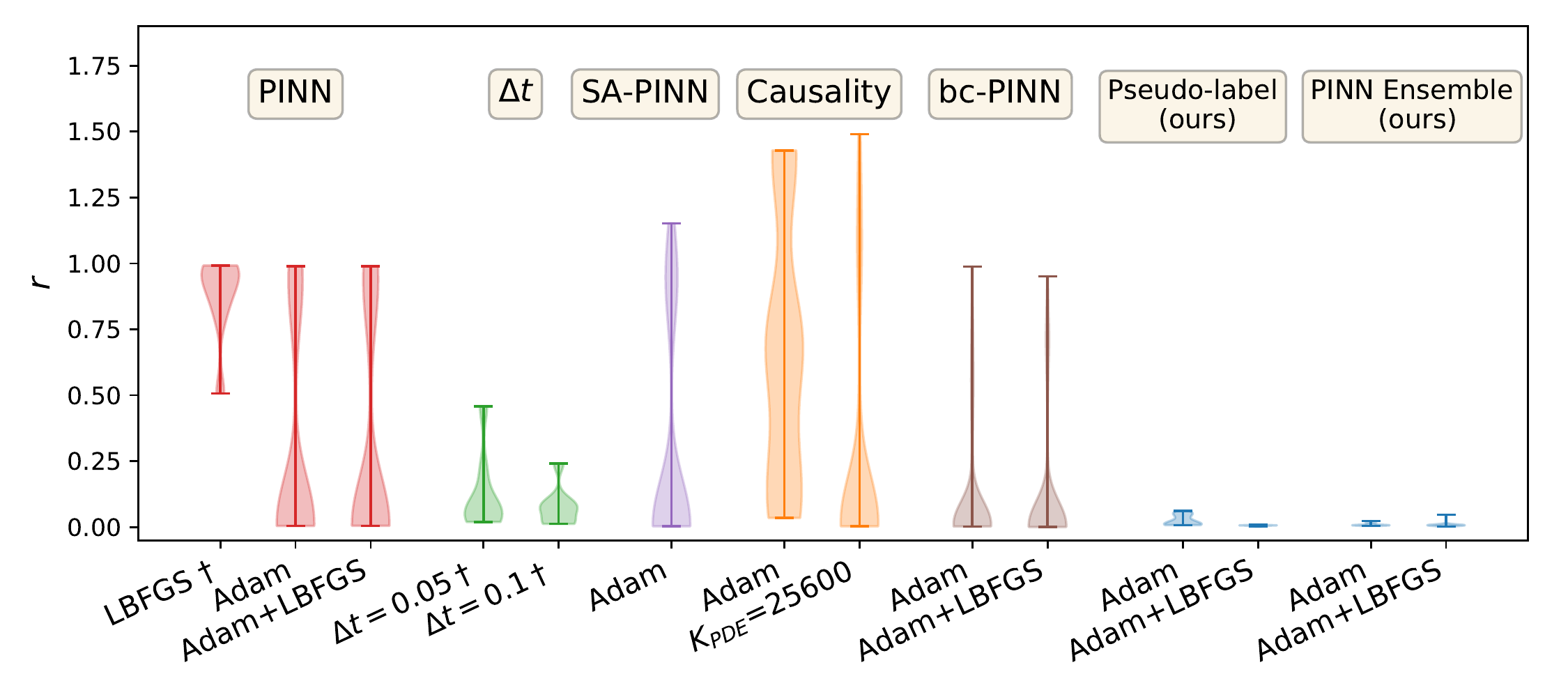}
  \end{center}
  \vskip -5mm
  \caption{The violin plots of the relative error $r$ for the proposed method (blue violins on the right) and strong PINN baselines in solving three systems (Eqs.~\ref{eq:conv}--\ref{eq:ibrd}). The proposed method yields stable training.}
\label{f:tab1_violin}
\end{wrapfigure}
In this work, we follow the idea of the gradual expansion of the solution interval when training PINNs. We propose to train an ensemble of PINNs and use the ensemble agreement (confidence) as the criterion for expanding the solution interval to new areas. Due to the flexibility of the domain expansion, in contrast to existing baselines, our algorithm can easily incorporate measurements in arbitrary locations: the algorithm will propagate the solution from all the locations where supervision targets are known (see Fig.~\ref{f:conv_02_ens} as an example). The proposed algorithm does not need a pre-defined schedule of interval expansion and it treats time and space equally. We experimentally show that the proposed algorithm can stabilize PINN training (see Fig.~\ref{f:tab1_violin}) and yield performance competitive to the recent variants of PINNs which use time adaptation.

\section{Background}

\subsection{Physics-informed neural networks}
\label{sec:pinn}

Physics-informed neural networks are neural networks which are trained to approximate the solution of a partial differential equation
\begin{equation}
    \dd{\U(x, t)}{t} = f \left(\dds{\U(x, t)}{x}, \dd{\U(x, t)}{x}, x, t \right)
\label{eq:pde}
\end{equation}
on the interval $x \in [X_1, X_2], t \in [T_1, T_2]$ with initial conditions
\begin{equation}
  \U(x, T_1) = \U_{0}(x), \quad x \in [X_1, X_2]
\label{eq:ic}
\end{equation}
and boundary conditions
\begin{equation}
\vg (\U(x, t)) = 0, \quad x \in \{X_1, X_2\} \,.
\label{eq:bc}
\end{equation}
Functions $f$ and $\U_{0}$ in Eqs.~\ref{eq:pde}--\ref{eq:ic} are assumed to be known, function
$\vg$ in Eq.~\ref{eq:bc} is known as well and it can represent different types of boundary conditions (e.g., Neumann, Dirichlet, Robin or periodic boundary conditions).
In the PINN approach, one approximates the solution with a neural network which takes inputs $x$ and $t$ and produces $\U(x, t)$ as the output.
The network is trained by minimizing a sum
of several terms:
\begin{equation}
    \mathcal{L} = \mathcal{L}_\text{S} + \mathcal{L}_\text{B} + \mathcal{L}_\text{PDE} \text{. }
\label{eq:loss}
\end{equation}
$\mathcal{L}_\text{S}$ is the standard supervised learning loss which makes the neural network fit the initial conditions:
\begin{align}
  \Lsup = \sum_{i=1}^{\Ks} w_i \left(\U(x_i, t_i) - \U_i \right)^2
  \,.
\label{eq:losssup}
\end{align}
where $w_i$ are point-specific weights,
$t_i = T_1$, $\U_i = \U_0(x_i)$ and $x_i$ are sampled from $[X_1, X_2]$.
$\mathcal{L}_\text{B}$ is the loss computed to satisfy the boundary conditions:
\begin{align}
  \Lb = \sum_{j=1}^{\Kb} w_j ||\vg (\U(x_j, t_j))||^2
  \,,
\label{eq:lossbcond}
\end{align}
where $w_j$ are point-specific weights, $x_j \in \{X_1, X_2\}$ and $t_j$ are sampled from $[T_1, T_2]$.
$\mathcal{L}_\text{PDE}$ is the loss derived from the PDE in~\ref{eq:pde}:
\begin{align}
  \mathcal{L}_\text{PDE} = \sum_{k=1}^{\Kpde}
  w_k \left(\dd{\U_k}{t} - f \left(\dds{\U_k}{x}, \dd{\U_k}{x}, x_k, t_k \right)\right)^2
\,,
\label{eq:losspde}
\end{align}
where $w_k$ are point-specific weights and the partial derivatives $\der{\U_k}{t}$, $\dds{\U_k}{x}$, $\dd{\U_k}{x}$ are computed at collocation points $(x_k, t_k)$ sampled from the interval $x_k \in [X_1, X_2], t_k \in [T_1, T_2]$.
Classical PINNs use shared weights $w_\text{S} = w_i, \forall i$, $w_\text{B} = w_j, \forall j$, $w_\text{PDE} = w_k, \forall k$. The values of the weights vary depending on the implementation and in the simplest case they are $w_\text{S}=1/\Ks$, $w_\text{B}=1/\Kb$, $w_\text{PDE}=1/\Kpde$.

The method can easily be extended to fit a sequence of observations  $\{((x^*_i, t^*_i), \U^*_i)\}_{i=1}^N$ by including the observed data to the supervision loss in Eq.~\ref{eq:losssup}.
In this case, the method can be seen as fitting a neural network to the training data (containing the observations) while regularizing the solution using the PDE loss in Eq.~\ref{eq:losspde}. One advantage of the PINN method compared to traditional numerical solvers is the ability to work with ill-posed problems, for example, if the initial conditions are known only in a subset of points. More details on existing extensions of PINNs can be found in Appendix~\ref{sec:pinn_extension}.

\subsection{PINNs with expansion of the time interval}

Recently, several papers have proposed to solve the initial-boundary value problem by gradually expanding the interval from which points $(x_k, t_k)$ in Eq.~\ref{eq:losspde} are sampled. These versions of PINNs are closest to our approach.

\paragraph{Time-adaptive strategies}

\citet{wight2020solving} 
and \citet{krishnapriyan2021characterizing} show the effectiveness of the time marching strategy: the time interval $[T_1, T_2]$ is split into multiple sub-intervals and the equation is solved on each individual sub-interval sequentially by a separate PINN. The solution at the border of the previous sub-interval is used as the initial condition for the next one.

A similar approach is based on progressive expansion of the time interval by gradually increasing the end point $T_2$ during training \citep{wight2020solving, mattey2022novel}.
Backward compatible PINNs \citep[bc-PINNs, ][]{mattey2022novel} implement this idea such that the solution found during the previous interval extension is used as the PINN targets during training on a newly expanded interval to prevent catastrophic forgetting.

All these time adaptation strategies need a pre-defined schedule for the time interval expansion, which makes them similar to classical numerical solvers which use pre-defined discretization schemes.

\paragraph{Causality training}

The idea of time adaptation is closely related to the adaptive weighting scheme that respects causality \citep{wang2022respecting}.
The authors use adaptive weights for the individual terms in Eq.~\ref{eq:losspde} such that the weights are computed using the cumulative PDE loss for the preceding points:
\begin{align}
\textstyle
w_k = \exp\left(
- \sum_{k' | t_{k'} < t_k} \epsilon \mathcal{L}_\text{PDE}(t_{k'})
\right)
,
\label{eq:causality}
\end{align}
where $\mathcal{L}_\text{PDE}(t_{k'})$ denotes an individual term in Eq.~\ref{eq:losspde} that corresponds to time point $t_{k'}$.
The idea is to zero out the effect of the points that are far away from the initial conditions until the solution is approximated well on all the points before them. The method assumes that the boundary conditions are enforced as hard constraints and thus the total loss consists of  $\mathcal{L}_\text{S}$ and $\mathcal{L}_\text{PDE}$.

\section{Ensembles of PINNs}
\subsection{Motivation}
\label{sec:failures}

\begin{figure*}[t]
\newcommand{\myheight}{36mm}
\begin{minipage}[b]{.31\linewidth}
\centering
\includegraphics[height=\myheight,trim={16mm 6mm 42mm 10mm},clip]{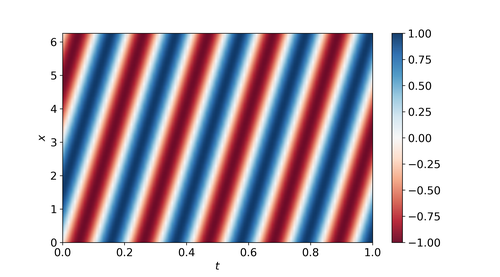}
\\[-1mm]
\centering
\small (a) ground truth solution
\end{minipage}
\begin{minipage}[b]{.31\linewidth}
\includegraphics[height=\myheight,trim={16mm 6mm 42mm 10mm},clip]{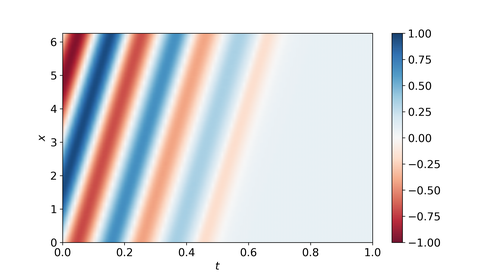}
\\[-1mm]
\centering
\small (b) PINN solution 1 %
\end{minipage}
\begin{minipage}[b]{.36\linewidth}
\includegraphics[height=\myheight,trim={16mm 6mm 21mm 10mm},clip]{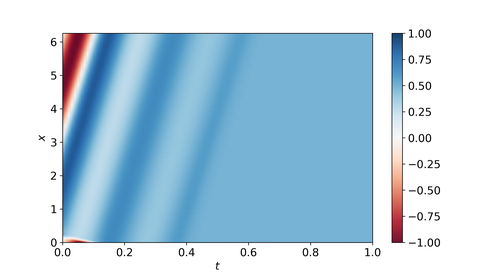}
\\[-1mm]
\centering
\small (c) PINN solution 2 %
\end{minipage}
\\[2mm]
\begin{minipage}[b]{.31\linewidth}
\hspace{.31\linewidth}
\end{minipage}
\begin{minipage}[b]{.31\linewidth}
\includegraphics[height=\myheight,trim={16mm 6mm 42mm 10mm},clip]{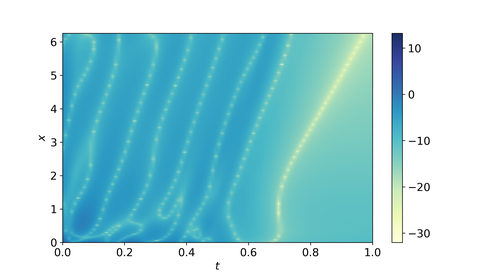}
\\[-1mm]
\centering
\small (d) PDE error for PINN solution 1
\end{minipage}
\begin{minipage}[b]{.36\linewidth}
\includegraphics[height=\myheight,trim={16mm 6mm 21mm 10mm},clip]{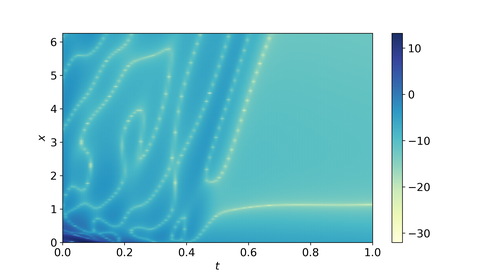}
\\[-1mm]
\centering
\small (e) PDE error for PINN solution 2
\end{minipage}

\caption{Example of solving the convection system (Eqs.~\ref{eq:conv}--\ref{eq:ibconv}) for $\beta=30$, $t \in [0, 1]$ using PINNs with the LFBGS optimizer. (a): The ground truth solution. (b)-(c): Two solutions found with different initializations of PINNs. (d)-(e): The logarithm of the individual terms in Eq.~\ref{eq:losspde} as a function of $x$ and $t$ for the solutions.}
\label{f:pinn_different_seeds}
\end{figure*}

To motivate our approach, we demonstrate failure cases of PINNs using an example from \citep{krishnapriyan2021characterizing} on solving a convection equation. Fig.~\ref{f:pinn_different_seeds} shows the ground-truth solution of the equation (Fig.~\ref{f:pinn_different_seeds}a) and two inaccurate solutions (Figs.~\ref{f:pinn_different_seeds}b, c) found by PINNs trained with the same network architecture but different random seeds for weight initialization. 
The found PINN solutions can be seen as a combination of two solutions: the correct solution near the initial conditions (for small $t$) and a simpler solution farther away from the initial conditions (for large $t$).
The second row  in Fig.~\ref{f:pinn_different_seeds} illustrates that the second, simpler solution satisfies well the solved PDEs.
This example illustrates that simple solutions can be attractive for PINNs, which can cause problems for the optimization procedure. Once a PINN finds a simple, locally consistent solution in some areas (for example, far away from the initial conditions), it may be very difficult to change it. This leads to a final solution which is a combination of the correct solution and a wrong one.

This example provides the following intuition: when points located far away from the initial conditions contribute to the PDE loss in Eq.~\ref{eq:losspde}, it may hurt the optimization procedure by pulling the solution towards a bad local optimum. On the other hand, including those points in the PDE loss at the beginning of training hardly brings any benefits: it makes little sense to regularize the solution using the loss in Eq.~\ref{eq:losspde} before we know its approximate shape. 
The ability to escape from wrong solutions largely depends on the design choices made for the PINN training such as the optimizer type,
the use of mini-batches, type of the sampling utilized for points in $\mathcal{L}_\text{PDE}$ term in Eq.~\ref{eq:losspde}, normalization of the inputs, hard or soft encoding of the initial and boundary conditions and so on. While some of these tricks can be beneficial for the PINN accuracy on particular systems, it is quite difficult to select a common set of settings beneficial across a wider range of PDEs.

The illustrated problem is avoided by the classical numerical solvers because they usually ``propagate'' the solution from the initial and boundary conditions to cover the entire interval using a schedule determined by the discretization scheme.

\subsection{Method}

In this paper, we propose to gradually expand the areas from which we sample collocation points $(x_k, t_k)$ to compute loss $\mathcal{L}_\text{PDE}$.
Our approach is based on training an ensemble of PINNs: a set of neural networks initialized with different weights but trained using the same loss function.
Since PINN ensemble members typically converge to the same solution in the vicinity of observed data but may favor distinct wrong solutions farther away from the observations  (Fig.~\ref{f:pinn_different_seeds}b-c), we can use the ensemble agreement as the criterion for including new points for computing loss $\mathcal{L}_\text{PDE}$. Optionally, if all ensemble members agree on the solution in a particular point, we can create a pseudo-label (taken as the median of the ensemble predictions) for that point and make this point contribute to the supervision loss $\mathcal{L}_\text{S}$ in Eq.~\ref{eq:losssup}. 

\begin{figure*}[th]
\newcommand{\myheight}{34mm}
\centering
\begin{tblr}{
  colspec = {X[0.83,c]X[c,h]X[c,h]},
  stretch = 0,
  rowsep = 0pt,
}
  At initialization & After 5000 updates & After 24000 updates
\\
  \includegraphics[height=\myheight,trim={16mm 6mm 42mm 10mm},clip]{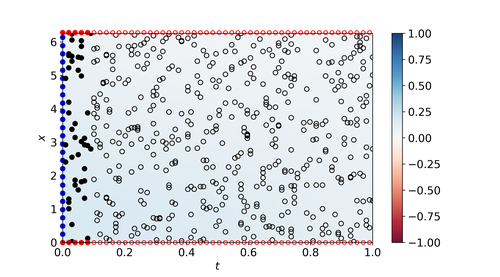} &
  \includegraphics[height=\myheight,trim={16mm 6mm 21mm 10mm},clip]{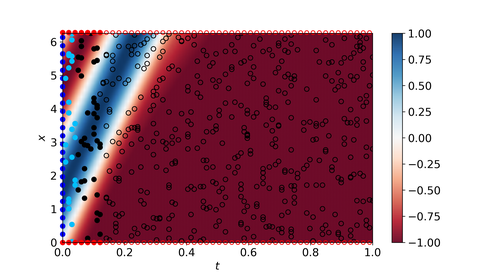} &
  \includegraphics[height=\myheight,trim={16mm 6mm 21mm 10mm},clip]{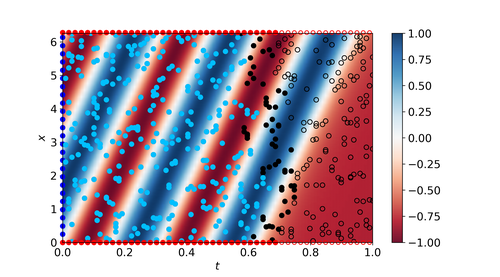}
\\
  & 
  \includegraphics[height=\myheight,trim={16mm 6mm 21mm 10mm},clip]{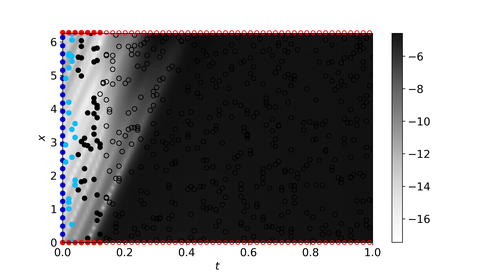} &
  \includegraphics[height=\myheight,trim={16mm 6mm 21mm 10mm},clip]{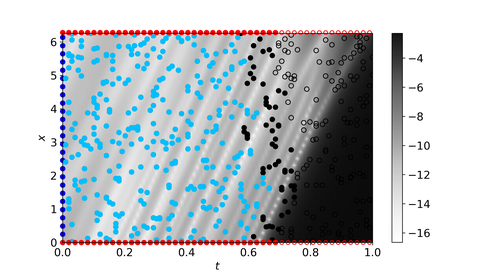}
\end{tblr}

\caption{Illustration of the proposed algorithm on solving the convection system with $\beta=20$ on $t \in[0, 1]$. Top row: ensemble median, bottom row: log-variance of ensemble predictions.
Dark blue dots: set $D_\text{L}$ of points which contribute to loss $\mathcal{L}_\text{S}$; %
light blue dots: set $D_\text{PL}$ of points with pseudo-labels (contribute either to $\mathcal{L}_\text{S}$ or $\mathcal{L}_\text{PDE}$); black dots: points that contribute to loss $\mathcal{L}_\text{PDE}$;
red dots: points that contribute to loss 
$\mathcal{L}_\text{B}$. %
}
\label{f:alg}
\end{figure*}

\begin{algorithm*}[pht]
\caption{Training PINNs with label propagation}
\label{alg:ensemble}
\begin{algorithmic}[1]
\Require $\Delta_\text{PDE}$, $\Delta$, $\sigma^2$ and $\epsilon$
\State $D_\text{L} = \{((x_i, t_i), \U_i)\}$ \Comment Points with targets (blue dots)
\State $D_\text{PL} = \{\}$ \Comment Points with pseudo-labels (light blue dots)
\State $I_\text{B} = \{(x_j, t_j)\}$ \Comment Sample candidate points for computing $\mathcal{L}_\text{B}$ (empty red circles)
\State $I_\text{PDE} = \{(x_m, t_m)\}$ \Comment Sample candidate points for computing $\mathcal{L}_\text{PDE}$ (empty black circles)

\While{not converge}

\State $D = D_\text{L} \cup D_\text{PL}$
\State $I_\text{B}' = \{(x_j, t_j) \in I_\text{B} \mid \Call{distance}{(x_j, t_j), D} < \Delta_\text{PDE}\}$  \Comment red dots
\State $I_\text{PDE}' = \{(x_m, t_m) \in I_\text{PDE} \mid \Call{distance}{(x_m, t_m), D} < \Delta_\text{PDE} \}$ \Comment black dots

\State Train $L$ networks $f_l$ for $N$ iterations
using $D_\text{L} \cup D_\text{PL}$ ('PL' version) or $D_\text{L} $ ('Ens' version), $I_\text{B}'$, $I_\text{PDE}'$ to compute $\mathcal{L}_\text{S}$, $\mathcal{L}_\text{B}$, $\mathcal{L}_\text{PDE}$, respectively.

\For{$(x_m, t_m) \in I_\text{PDE}'$}
    \State $\hat{\U}_{l} = f_l(x_m, t_m), \quad \forall l \in 1, ..., L$ \Comment Prediction of each ensemble network
    \State $v_m = \text{variance}(\hat{\U}_1, \dots, \hat{\U}_L)$ \Comment Variance of predictions
    \State ${\bar{\U}}_m = \text{median}(\hat{\U}_1, \dots, \hat{\U}_L)$ \Comment Median of predictions
    \If{$v_m < \sigma^2 \And \Call{distance}{(x_m, t_m), D} < \Delta$}
        \State $D_\text{PL} \gets D_\text{PL} \cup ((x_m, t_m), {\bar{\U}}_m)$ \Comment Add a point with a pseudo-label
    \EndIf
    \EndFor

\EndWhile
\\
\Function{distance}{$(x, t), D$}
    \State $D' = \{(x_i, t_i) \in D \mid || \frac{1}{L} \sum_l f_l(x_i, t_i) - \U_i || < \epsilon \}$  \Comment Points with good fit to targets
    \State \Return $\min_{(x_i, t_i) \in D'} || (x_i, t_i) - (x, t) ||$
\EndFunction
\end{algorithmic}
\end{algorithm*}

The proposed algorithm is illustrated in Fig.~\ref{f:alg}. At the beginning of training, the supervision loss $\mathcal{L}_\text{S}$ is computed using points sampled at the initial conditions (the blue dots in Fig.~\ref{f:alg}a) and losses $\mathcal{L}_\text{PDE}$ and $\mathcal{L}_\text{B}$ are computed using only points that are close enough to the initial conditions (the black and red dots in Fig.~\ref{f:alg}a respectively). The proximity is measured by thresholding the Euclidean distance to the closest point among the blue dots. After $N$ training iterations, we compute the median and the variance of the ensemble predictions (see Fig.~\ref{f:alg}b-c). If the variance in a particular location is small enough, we use the median of the ensemble predictions at that point as a pseudo-label and add that point to the data set which is used to compute the supervision loss $\mathcal{L}_\text{S}$ (the light blue dots in Fig.~\ref{f:alg}b).
Points for pseudo-labeling are selected among the collocation points (the black dots in Fig.~\ref{f:alg}a).
Locations which are close enough to the data points with labels or pseudo-labels are added to the set which contributes to $\mathcal{L}_\text{PDE}$ and $\mathcal{L}_\text{B}$ (the black and red dots in Fig.~\ref{f:alg}b). Then, we train the ensemble of PINNs for a fixed number of iterations and again increase the sets of inputs which are used to compute the losses in a similar way (Fig.~\ref{f:alg}d-e). The iterations continue until all collocation points (which are pre-sampled at the beginning of the training procedure) are included in the loss computations. 

More formally, the training procedure is presented in Algorithm~\ref{alg:ensemble}. 
The dots and the circles in the algorithm refer to Fig.~\ref{f:alg}. 
In the experiments, we test two versions of the proposed algorithm:
\begin{enumerate}
    \item \textit{Pseudo-labels (PL):} a version with pseudo-labels in which the points with a high degree of ensemble agreement are used to compute both $\Lpde$ and $\Lsup$;
    \item \textit{PINN Ensemble (Ens):} a version without pseudo-labels, in which the points with a high degree of ensemble agreement are used to compute $\Lpde$ but not $\Lsup$.
\end{enumerate}

Each member of the ensemble is trained to minimize the loss in Eq.~\ref{eq:loss} with shared weights $w_\text{B}=1/|I_\text{B}|$, $w_\text{PDE}=1/|I_\text{PDE}|$ and $w_\text{S}=1/|D|$ for the PINN Ensemble version and $w_\text{S}=1/|D \cup I_\text{PDE}|$ for the version with pseudo-labels.
Other weighting strategies \citep[e.g.,][]{wight2020solving,wang2021understanding,wang2021ntk} could be combined with our approach as well.

\subsection{Related work}
\paragraph{PINNs and their extensions}

The proposed algorithm is built on the idea of the gradual expansion of the solution interval, which makes it similar to the time-adaptive techniques \citep{wight2020solving,krishnapriyan2021characterizing, mattey2022novel} and the adaptive weighting method that respects causality \citep{wang2022respecting}.
The advantage of the proposed algorithm is its greater flexibility in the way of expanding the area covered by collocation points: instead of expanding the time interval with a pre-defined \citep{wight2020solving,krishnapriyan2021characterizing, mattey2022novel} or an automatic \citep{wang2022respecting} schedule, our algorithm considers each collocation point individually during the expansion and it treats time and space equally.
This feature allows the application of the algorithm to datasets with an arbitrary layout of the points with known targets. We illustrate this by solving the convection system (Eqs.~\ref{eq:conv}--~\ref{eq:ibconv}) on the interval  $t \in [0, 2]$ when the solution is known for $t=0$ and $t=2$. As illustrated in Fig.~\ref{f:conv_02_ens}, the algorithm finds a reasonable schedule for expanding the area starting from both ends of the interval.

\paragraph{Label propagation and ensembles}
Training an ensemble of PINNs with pseudo-labeling is related to how label propagation is done in semi-supervised classification tasks \citep[see, e.g,][]{lee2013pseudo,sohn2020fixmatch}: when a classifier becomes confident in the predicted class of an unlabeled example, that example is added to the labeled set. Model ensembles~\citep{laine2017temporal} or prediction ensembles~\citep{berthelot2019mixmatch} are often used in those tasks to generate better targets. Since PINNs are trained on real-valued targets, one can view PINNs as regression models regularized by the loss in Eq.~\ref{eq:losspde}.
Label propagation in regression tasks is less studied with only a few existing works on semi-supervised regression~\citep{jean2018ssdkl,li2017learning}. In our algorithm, we use the confidence of the ensemble predictions to decide whether the solution interval can be extended and which points can be assigned pseudo-labels.

\section{Experiments}

We test the proposed algorithm on finding the solutions of the following differential equations:
\begin{itemize}[wide]
    \item convection equation used to model transport phenomena 
    \begin{align}
        \label{eq:conv}
        \dd{u}{t} =& -\beta \dd{u}{x} \text{,} \quad x \in [0, 2\pi] \text{,} \quad t \in [0, 1] \text{,} \quad \beta = \text{const} \\
        \label{eq:ibconv}
        u(x, 0) =& \sin(x)\text{,} \quad u(0, t) = u(2\pi,t)
    \end{align}
    
    \item reaction system for modelling chemical reactions
    \begin{align}
        \label{eq:react}
        \dd{u}{t} =& \rho u (1-u) \text{,} \quad x \in [0, 2\pi] \text{,} \quad t \in [0, 1] \text{,} \quad \rho = \text{const} \\
        \label{eq:ibreact}
        u(x, 0) =& \exp\left(-{8(x - \pi)^2}/{\pi^2} \right) \text{,} \quad u(0, t) = u(2\pi,t)
    \end{align}
    
    \item reaction-diffusion equation that models reactions together with diffusion of substances 
    \begin{align}
        \label{eq:rd}
        \dd{u}{t} =& \nu \dds{u}{x} + \rho u (1-u) \text{,} \quad x \in [0, 2\pi] \text{,} \quad t \in [0, 1] \text{,} \quad \nu\text{, } \rho = \text{const} \text{,} \\
        u(x, 0) =& \exp\left(-{8(x - \pi)^2}/{\pi^2} \right) \text{,} \quad
        u(0, t) = u(2\pi,t) \text{,} \quad u_x\left(0, t \right) = u_x \left(2\pi,t \right)
        \label{eq:ibrd}
    \end{align}
    
    \item diffusion equation with periodic boundary conditions
    \begin{align}
        \label{eq:diff}
        \dd{u}{t} &= \frac{1}{d^2} \dds{u}{x} \text{,} \quad x \in [0, 2\pi] \text{, } \quad t \in [0, 1] \text{,} \quad d = \text{const} \\ 
        \label{eq:ibdiff}
        u(x, 0) &= \sin(d x) \quad u(0, t) = u(2\pi,t) \text{,} \quad u_x\left(0, t \right) = u_x \left(2\pi,t \right)
    \end{align}
    
    \item diffusion equation in~\ref{eq:diff} with boundary conditions of the Dirichlet type:
    \begin{align}
        \label{eq:ibdiff_dir}
        u(0, t) = u(2\pi,t) = 0
    \end{align}
\end{itemize}

\begin{figure*}[t]
\newcommand{\myheight}{35mm}
\centering

\begin{minipage}{0.25\linewidth}
\centering
\small Start of training
\end{minipage}
\begin{minipage}{0.3\linewidth}
\centering
\small Intermediate results
\end{minipage}
\begin{minipage}{0.25\linewidth}
\centering
\hspace{-6mm}
\small Final results
\end{minipage}
\\[1mm]
\includegraphics[height=\myheight,trim={16mm 7mm 42mm 10mm},clip]{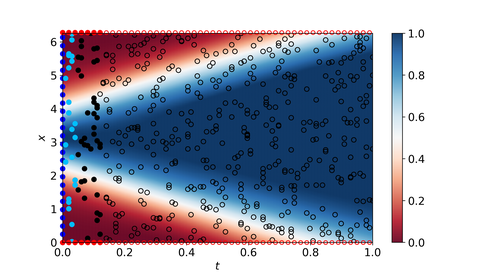}
\includegraphics[height=\myheight,trim={16mm 7mm 42mm 10mm},clip]{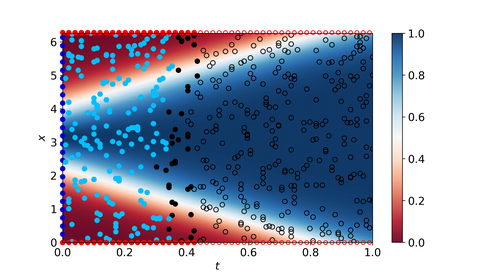}
\includegraphics[height=\myheight,trim={16mm 7mm 21mm 10mm},clip]{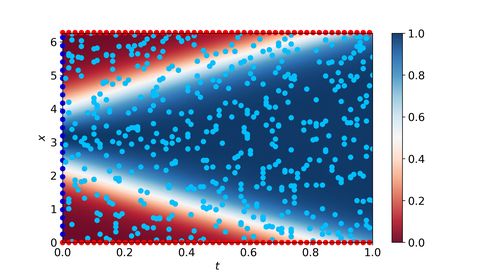}
\\ %
\centering
\small (a) reaction equation (Eqs.~\ref{eq:react}--\ref{eq:ibreact}), $\rho=8$
\\[2mm]
\includegraphics[height=\myheight,trim={16mm 7mm 42mm 10mm},clip]{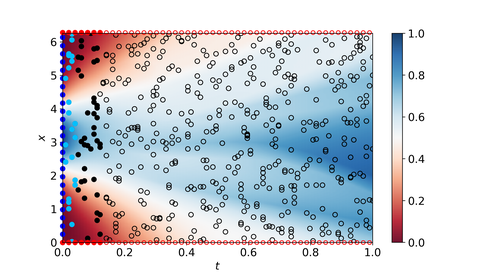}
\includegraphics[height=\myheight,trim={16mm 7mm 42mm 10mm},clip]{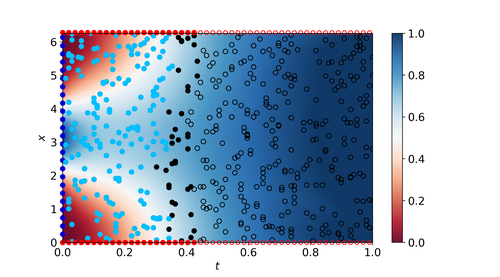}
\includegraphics[height=\myheight,trim={16mm 7mm 21mm 10mm},clip]{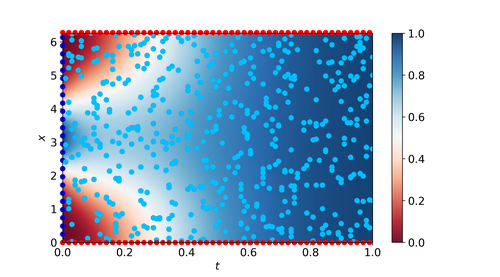}
\\
\small (b) reaction-diffusion equation (Eqs.~\ref{eq:rd}--\ref{eq:ibrd}),
$\rho=5$, $\nu=5$
\\[2mm]
\includegraphics[height=\myheight,trim={16mm 7mm 42mm 10mm},clip]{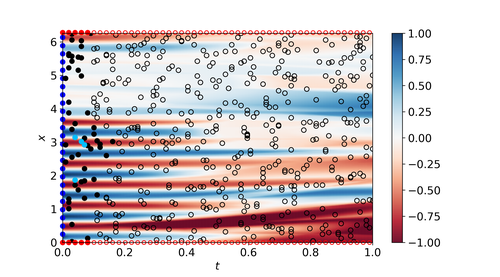}
\includegraphics[height=\myheight,trim={16mm 7mm 42mm 10mm},clip]{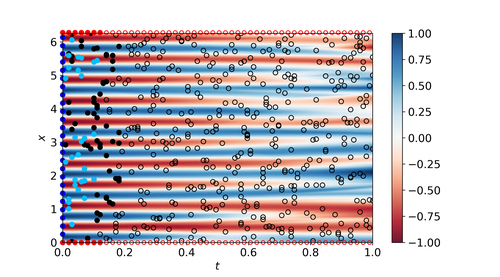}
\includegraphics[height=\myheight,trim={16mm 7mm 21mm 10mm},clip]{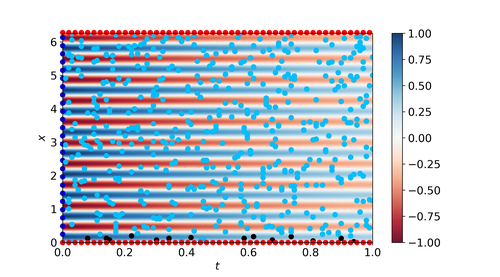}
\\ %
\centering
\small (c) diffusion equation
with periodic boundary conditions (Eqs.~\ref{eq:diff}--\ref{eq:ibdiff}), $d=10$

\caption{The training progress of PINN ensembles with pseudo-labels.
The dot color scheme is from Fig.~\ref{f:alg}.
}
\label{f:diff_rd_ens}
\end{figure*}

In all the experiments, we use a multi-layer perceptron with four hidden layers with 50 neurons and the tanh activation in each hidden layer as a backbone PINN. Our ensemble of PINNs contains five such networks. The two inputs $x$ and $t$ of the network are normalized to $[-1, 1]$, which has a positive effect on the accuracy in our experiments. The models are trained either with the Adam optimizer \citep{kingma2017adam} with  learning rate 0.001 or with Adam followed by fine-tuning with LBFGS \citep[][]{liu1989limited}. The LBFGS fine-tuning is done before each update of the collocation points which contribute to loss $\Lpde$ and at the very end of training.

Illustrations of the training procedure for the considered systems can be found in Figs.~\ref{f:alg} and \ref{f:diff_rd_ens}. The plots show that the proposed algorithm finds accurate solutions for the considered systems.

\begin{table*}[t]
\caption{
The means (and standard deviations) across 10 random seeds of the relative $l_2$ errors in Eq.~\ref{eq:r} for the convection (Eqs.~\ref{eq:conv}--\ref{eq:ibconv}), reaction (Eqs.~\ref{eq:react}--\ref{eq:ibreact}) and reaction-diffusion (Eqs.~\ref{eq:rd}--\ref{eq:ibrd}) systems. All numbers should be multiplied by $10^{-3}$.
The rows marked with $^*$ show the results obtained with $\Kpde=25600$ collocation points, otherwise $\Kpde=1000$.
In \textBF{bold}: the best three results.
}
\label{tab:results}
\begin{center}
\begin{tabular}{@{}c|cc|cc|cc}
\multirow{2}{*}{Method} & \multicolumn{2}{c|}{Convection, $\beta$} & \multicolumn{2}{c|}{Reaction, $\rho$} & \multicolumn{2}{c}{Reaction-diffusion, $\nu$}
\\
& $\beta=30$ & $\beta=40$ & $\rho=5$ & $\rho=7$ & $\nu=3$ & $\nu=4$
\\
\hline
PINN, LBFGS %
          & 827 (39) & 887 (21)
          & 985 (7) & 997 (2)
          & 726 (346) & 799 (257)
\\
PINN, Adam
          & 13.4 (3.2) & 16.2 (4.1)
          & 11.9 (6.2) & 887 (269)
          & 252 (375) & 437 (433)
\\
PINN, Adam+LBFGS
          & \textBF{9.02 (2.03)} & \textBF{13.1 (2.7)}
          & 10.8 (3.5) & 967 (26)
          & 170 (328) & 426 (427)
\\
SA-PINN %
        & 925 (35) & 980 (58)
        & \textBF{4.48 (0.94)} & 11.5 (2.3)
        & \textBF{5.11 (0.53)} & 5.94 (1.63)
\\
Causality
          & 1425 (4) & 1393 (3)
          & 41.4 (4.2) & 220 (23)
         & 700 (4) & 713 (3)
\\
Causality$^*$ %
          & 436 (496) & 1104 (228)
          & 6.01 (1.87) & 43.8 (8.0)
          & 6.51 (0.24) & 6.56 (0.17)
\\
bc-PINN, Adam
        & 212 (251) & 405 (331)
        & 5.37 (2.69) & \textBF{8.45 (2.14)}
        & 6.80 (2.70) & 8.94 (4.44)
\\
bc-PINN, Adam+LBFGS
        & 21.2 (40.8) & 580 (301)
        & \textBF{1.86 (0.81)} & \textBF{3.23 (2.99)}
        & 6.65 (0.11) & \textBF{6.51 (0.13)}
\\
\hline
\hline
PL, Adam (our)
          & 21.3 (1.5) & 51.0 (4.8)
          & 21.3 (2.5) & 55.7 (4.1)
          & 8.35 (0.42) & 7.95 (0.47)
\\
PL, Adam+LBFGS (our)
          & \textBF{5.42 (1.23)} & \textBF{6.72 (1.37)}
          & \textBF{2.41 (0.44)} & \textBF{6.02 (0.85)}
          & \textBF{6.59 (0.01)} & \textBF{6.45 (0.03)}
\\
Ens, Adam (our) %
          & \textBF{7.00 (2.74)} & \textBF{12.6 (3.0)}
          & 7.21 (1.57) & 13.4 (3.9)
          & 6.68 (0.37) & 6.63 (0.30)
          
\\
Ens, Adam+LBFGS (our) %
          & 8.69 (10.5) & 14.0 (11.8)
          & 6.83 (1.34) & 7.07 (2.14)
          & \textBF{6.62 (0.01)} & \textBF{6.48 (0.02)}
\\
\end{tabular}
\end{center}
\end{table*}

\begin{table*}[t]
\caption{
The means (and standard deviations) across 10 random seeds of the relative $l_2$ errors in Eq.~\ref{eq:r} for the diffusion system (Eq.~\ref{eq:diff}) with periodic and Dirichlet (Eqs.~\ref{eq:ibdiff}--~\ref{eq:ibdiff_dir}) boundary conditions.
All numbers should be multiplied by $10^{-3}$. The rows marked with $^*$ show the results obtained with $\Kpde=20000$, otherwise $\Kpde=1000$. In \textBF{bold}: the best three results.
}
\label{tab:diff_bc}
\begin{center}
\begin{tabular}{@{}c|ccc|ccc@{}}
\multirow{2}{*}{Method} & \multicolumn{3}{c|}{periodic boundary conditions} & \multicolumn{3}{c}{Dirichlet boundary conditions}
\\
\cline{2-7}
& $d=5$ & $d=7$ & $d=10$ & $d=5$ & $d=7$ & $d=10$
\\
\hline
PINN, LBFGS
          & 997 (.1) & 999 (.1) & 999 (.1)
          & 998 (.1) & 999 (.1) & 999 (.1)
\\
PINN, Adam 
          & 7.16 (5.61) & 20.1 (18.2) & 36.6 (18.1)
          & 10.6 (12.0) & 15.4 (14.8) & 17.9 (10.8)
          
\\
PINN, Adam+LBFGS
          & \textBF{0.30 (0.04)} & \textBF{0.60 (0.42)} & \textBF{1.00 (0.27)}
          & \textBF{0.32 (0.07)} & \textBF{0.39 (0.08)} & \textBF{0.54 (0.08)}
\\
SA-PINN, Adam %
        & 5.28 (1.68) & 5.94 (1.00) & 9.52 (2.61)
        & 4.85 (1.38) & 6.76 (1.46) & 8.08 (2.69)
        
\\
SA-PINN, Adam+LBFGS %
          & 2.59 (1.07) & 3.10 (1.62) & 6.46 (3.29)
          & 2.71 (0.96) & 3.53 (1.28) & 5.68 (2.85)

\\
bc-PINN, Adam %
        & 16.4 (6.1) & 82.2 (176) & 163 (212)
        & 13.6 (6.1) & 22.0 (7.6) & 50.4 (50.7)
        
\\
bc-PINN$^*$, Adam %
        & 10.7 (5.1) & 18.7 (6.9) & 48.1 (38.1)
        & 16.9 (12.2) & 20.5 (8.8) & 29.7 (16.1)
\\
bc-PINN, Adam+LBFGS
          & 4.49 (4.67) & 70.0 (190) & 108 (257)
          & 3.77 (2.87) & 6.14 (2.44) & 28.1 (44.2)
\\
\hline
\hline
PL, Adam (our)
        & 8.26 (3.52) & 11.5 (3.4) & 27.0 (8.0)
        & 9.15 (3.59) & 10.3 (4.0) & 25.9 (13.5)
\\
Ens, Adam (our)
        & 5.67 (3.02) & 10.1 (4.0) & 18.2 (6.7)
        & 8.07 (4.95) & 10.1 (5.2) & 14.5 (7.3)
\\ 
PL, Adam+LBFGS (our)
          & \textBF{0.30 (0.05)} & \textBF{0.53 (0.04)} & \textBF{1.64 (0.59)}
          & \textBF{0.29 (0.06)} & \textBF{0.55 (0.07)} & \textBF{1.73 (0.58)}
\\
Ens, Adam+LBFGS (our)
          & \textBF{0.12 (0.02)} & \textBF{0.41 (0.13)} & \textBF{1.59 (0.43)}
          & \textBF{0.11 (0.01)} & \textBF{0.33 (0.14)} & \textBF{2.03 (1.25)}
\end{tabular} 
\end{center}
\end{table*}

To evaluate the accuracy of the proposed algorithm, we compute metrics used in the previous works \citep{krishnapriyan2021characterizing,wang2022respecting}: we report the relative $l_2$ error
\begin{align}
\textstyle
    r = l_2(\hat{\vu} - \vu)/l_2(\vu)
\,,
\label{eq:r}
\end{align}
where $\vu$ is a vector of the ground-truth values in the test set, $\hat{\vu}$ is the corresponding PINN predictions and $l_2()$ denotes $l_2$ norm. Our test set consists of points on a regular $256 \times 100$ grid.

In Table~\ref{tab:results}, we compare the accuracy of the proposed method with several strong baselines proposed recently in the literature. We present the means and standard deviations of the solution errors obtained in 10 runs with different initializations. Since most of the considered methods benefit from longer training, we limit the number of network updates for all the comparison methods.
We use the following comparison methods:
(1) Vanilla PINN trained with the LBFGS optimizer, (2) Vanilla PINN trained with the Adam optimizer, 
(3) Causality \citep{wang2022respecting}, (4) SA-PINN \citep{mcclenny2020self}, (5) bc-PINN \citep{mattey2022novel}.
More details on the methods and the selection of hyperparameters can be found in Appendix~\ref{sec:train_details}.

The results in Table~\ref{tab:results} show that vanilla PINNs trained with LBFGS struggle to find accurate solutions for all the considered systems. Vanilla PINNs trained with Adam yield more accurate results but the results are unsatisfactory for the reaction and reaction-diffusion systems. The causality weighting scheme seems to require many more collocation points to find satisfactory solutions. SA-PINN and bc-PINN work very well on the reaction and reaction-diffusion systems but do not find good solutions for the convection system.
The proposed ensemble method provides stable training (see the summary of the results in Fig.~\ref{f:tab1_violin}) and shows competitive performance in all the considered systems.
We can also observe that the LBFGS fine-tuning has positive effect on the accuracy when a good approximation is found during pre-training with Adam.

In Table~\ref{tab:diff_bc}, we compare the accuracy of the proposed algorithm on solving the diffusion system with the two types of boundary conditions: periodic (Eq.~\ref{eq:ibdiff}) and Dirichlet-type (Eq.~\ref{eq:ibdiff_dir}) boundary conditions. In this comparison, we omit the causality-motivated baseline~\citep{wang2022respecting} because the periodic boundary conditions in the existing implementations are enforced as hard constraints. The results in Table~\ref{tab:diff_bc} show that training ensembles of PINNs yields competitive performance in these settings as well.

In Appendix~\ref{sec:hyperparam_sensi}, we study the sensitivity of the ensemble training of PINNs to its hyperparameters. The results show that the PINN Ensemble algorithm is generally stable at solving the considered systems with little sensitivity to the hyperparameter values.
We note the importance of hyperparameter $\Delta_\text{PDE}$ which determines how far the points considered for inclusion can be from the points already included in the loss calculations. The performance of the algorithm drops when $\Delta_\text{PDE}$ is too large. This happens because ensemble members may be attracted by the same trivial solution in areas too far away from the initial conditions, the effect caused by commonly used network architectures and initialization schemes \citep[see, e.g.,][]{wong2022learning, rohrhofer2022understanding}, which may negatively affect the ensemble diversity.

\section{Discussion and future work}

In this paper, we propose to stabilize training of PINNs by gradual expansion of the solution interval based on the agreement of an ensemble of PINNs. The obtained results suggest that the proposed approach can reduce the number of failure cases during PINN training. Another potential advantage of the proposed algorithm is that the PINN ensemble produces confidence intervals which can be viewed as uncertainty estimates of the found solution (see Fig.~\ref{f:alg}c, e). Although the proposed algorithm is more computationally expensive compared to vanilla PINNs, ensemble training can be effectively parallelized in which case the wall clock time of training does not grow significantly. The method shows good results for simple systems such as convection and reaction-diffusion.

This work can be extended in a number of ways. One potential direction is to use different ways of creating model ensembles, for example, by dropout~\citep{hinton2012dropout} or pseudo-ensembles~\citep{bachman2014learning}. 
It is interesting to investigate how the proposed algorithm can be combined with other tricks from the PINN literature, for example, adaptive balancing of the loss terms \citep{wang2021ntk}. %
Another direction is to find alternative ways of the solution interval expansion (e.g., update sets $\mathcal{I}_\text{PDE}'$, $\mathcal{I}_\text{B}'$ more frequently), which may increase the convergence speed.
It should also be possible to improve the PINN architecture such that the knowledge of a found (local) solution in one area could be used in finding the solution in another area. Using neural networks with the right inductive bias \citep[e.g., similar to the ones proposed by][]{sanchez2020learning,iakovlev2021learning,brandstetter2022message}, might provide a solution for that.

\section*{Acknowledgment}
We thank CSC (IT Center for Science, Finland) for computational resources and the Academy of Finland for the support within the Flagship programme: Finnish Center for Artificial Intelligence (FCAI).

\bibliography{references}

\appendix
\section{Appendix}
\subsection{PINN extensions}
\label{sec:pinn_extension}

Despite the elegance of the PINN approach, 
the method is known to be prone to failures, especially when the solution has a complex shape on the considered interval \citep[see, e.g., ][]{wang2021understanding,krishnapriyan2021characterizing,wang2021ntk, wang2021eigenvector, wang2022respecting, wight2020solving}. The optimization problem solved while training PINNs is very hard and the accuracy of the found solution is very sensitive to the hyperparameters of PINNs \citep[see, e.g., ][]{markidis2021old}: the weights $w_\text{S}$, $w_\text{B}$, $w_\text{PDE}$ of the loss terms, the parameterization used in the neural network and the strategy for sampling the points in which the loss terms are computed. 

Balancing shared weights $w_\text{S}$, $w_\text{B}$, $w_\text{PDE}$ of the individual loss terms in Eqs.~\ref{eq:losssup}--\ref{eq:losspde} is a popular way to improve the accuracy of the PINN solution. \citet{wight2020solving} propose using larger weights $w_\text{S}$ relative to $w_\text{B}$ and $w_\text{PDE}$ because the initial conditions largely determine the shape of the solution. Several works propose different schemes for dynamically adapting $w_\text{S}$, $w_\text{B}$, $w_\text{PDE}$ during training such that the weights that correspond to problematic loss terms get higher values. \citet{wang2021understanding,wang2021ntk} propose to adjust the weights either based on the magnitudes of the gradients of the corresponding loss terms  or based on the eigenvalues of the limiting Neural Tangent Kernel. \citet{liu2021dual} propose to treat weights $w_\text{S}$, $w_\text{B}$, $w_\text{PDE}$ as trainable parameters and adjust them jointly with the PINN parameters solving a minimax optimization problem.
Self-Adaptive PINNs \citep[SA-PINNs, ][]{mcclenny2020self} increase \emph{point-wise} weights $w_i$, $w_j$ and $w_k$ of the loss terms in Eqs.~\ref{eq:losssup}--\ref{eq:losspde} during training such that the changes of the weights are proportional to the corresponding loss terms.

The accuracy of PINNs can also be improved by using a different parameterization for the trained neural network instead of the most standard multilayer perceptron architecture. Several works  \citep[see, e.g.,][]{lagaris1998artificial,dong2021method, lu2021physics,sukumar2022exact} propose neural network parameterizations which guarantee that the initial or boundary conditions are satisfied exactly, thus eliminating terms $\mathcal{L}_\text{S}$ and $\mathcal{L}_\text{B}$ from Eq.~\ref{eq:loss}.
\citet{wang2021eigenvector} propose to use Fourier features as the inputs of the neural network, which is motivated by the success of Fourier feature networks \citep{tancik2020ffn} in preserving high-frequencies in the modeled solution. The method called SIREN \citep{sitzmann2019siren} proposes to use a multilayer perceptron with periodic activation functions and adjusts the weight initialization schemes to work better with such activation functions. This parameterization can also improve learning of high frequencies in the modeled solution. \citet{wang2021understanding} propose to add multiplicative connections to the standard multilayer perceptron architecture to account for possible multiplicative interactions between different input dimensions.

Another popular way of improving the accuracy of PINNs is to use adaptive strategies for sampling collocation points $(x_k, t_k)$ to compute $\mathcal{L}_\text{PDE}$ in Eq.~\ref{eq:losspde}.
\citet{wight2020solving} propose to sample more collocation points in the areas with large values of $\mathcal{L}_\text{PDE}$, which helps to learn solutions with fast transitions.
\citet{daw2022rethinking} propose an evolutionary strategy for sampling the collocation points: the points with large contribution to $\mathcal{L}_\text{PDE}$ are kept for the next iteration while the rest of the points are re-sampled uniformly from the domain.
These approaches resemble the strategy of the classical solvers to reduce the discretization interval when the solution cannot be estimated accurately.
More details on the sampling strategies for PINN and empirical comparison can be found in \citep{wu2022comprehensive}.

Many of the improvements proposed to PINNs can be combined, which is supported by existing software libraries \citep{lu2021deepxde,zubov2021neuralpde, hennigh2021nvidia}.

\subsection{Training details}
\label{sec:train_details}
For all comparison methods (similarly to \citet{krishnapriyan2021characterizing}), we use $\Kpde=1000$ collocation points randomly sampled from a regular $256 \times 100$ grid on the solution interval.
We use $\Ks=256$ points to fit the initial conditions, these points are selected from a regular grid on $x \in [0, 2\pi]$ with $t=0$. We use
$\Kb=100$ for the boundary condition loss $\mathcal{L}_\text{BC}$. These points are selected form a regular grid on $t \in [0, 1]$.

For the proposed ensemble method, we use the following hyperparameters: $\sigma^2 = 0.0004$, $\epsilon=0.001$, $\Delta = 0.05$ and $\Delta_\text{PDE}=0.1$. We perform $N=1000$ gradient updates before extending sets $I_\text{PL}'$, $I_\text{PDE}'$, $I_\text{B}$, except for the first set extension which is done after $N_1 = 5000$ training steps. We also set weighting coefficient $w_s = 64/|D \cup I_\text{PDE}|$ for runs with pseudo-labels trained with Adam and LBFGS similar to the closest baseline. 

For the baseline methods we adapted the authors' implementations with the following adjustments.
\begin{itemize}
    \item \textit{Causality \citep{wang2022respecting}:} PINN with an adaptive weighting scheme that respects causality.
    We report results for the values of hyperparameter $\epsilon$ in Eq.~\ref{eq:causality} that worked best for the considered systems: $\epsilon = 0.01$ for the convection system, $\epsilon = 100$ for the reaction system and $\epsilon = 1$ for the reaction-diffusion systems. We train the model until all the adapted weights become greater than 0.99 (the stopping criterion used by \citet{wang2022respecting}). As the existing implementation requires a regular grid, we use a grid of $\Kpde = 32 \times 32 = 1024$ collocation points to compute loss $\mathcal{L}_\text{PDE}$.
    We also report results with a denser grid of $\Kpde=256\times 100 = 25600$ points for this algorithm. We do not add normalization for network inputs $x$ and $t$ to $[-1, 1]$ and follow input encoding of the original implementation.
    
    \item \textit{SA-PINN \citep{mcclenny2020self}} with trainable point-specific weights. We
    switch off finetuning with the LBFGS optimizer after Adam due to observed training instabilities. We also add normalization of network inputs to $[-1, 1]$ similar to our method.
    
    \item \textit{bc-PINN \citep{mattey2022novel}}: a technique of expanding the time interval with a pre-defined schedule.
    We expand the time interval four times and sample 250 collocation points on each interval to compute loss $\mathcal{L}_\text{PDE}$ and 25 points per interval to compute $\mathcal{L}_\text{B}$. We report the results obtained with the Adam optimizer and with the Adam optimizer followed by LBFGS on each time interval.
    
\end{itemize}

For all of the methods, we train convection equation (Eqs.~\ref{eq:conv}--\ref{eq:ibconv}) with $\beta=30$ for 104000 gradient updates and with $\beta=40$ for 154000 updates, reaction (Eqs.~\ref{eq:react}--\ref{eq:ibreact}) and reaction-diffusion (Eqs.~\ref{eq:rd}--\ref{eq:ibrd}) equations are trained for 64000 updates, diffusion (Eqs.~\ref{eq:diff}--\ref{eq:ibdiff_dir}) equation with $d=5$ and $d=7$ for 84000 updates and with $d=10$ for 104000 (unless an automatic stopping criterion is used).

\subsection{Hyperparameter sensitivity}
\label{sec:hyperparam_sensi}

In Table~\ref{tab:ens_hyperparams}, we test the sensitivity of the ensemble training of PINNs to its hyperparameters. In these experiments, we use the number of updates as described in Appendix~\ref{sec:train_details}. We consider that a run has not converged if less than 95\% of the sampled points have been added to the set which is used to compute $\Lpde$ by the last update. Such runs are excluded from the statistics reported in Table~\ref{tab:ens_hyperparams}. This step is done only in the hyperparameter sensitivity experiments.

The most important hyperparameters are the variance threshold $\sigma^2$ and the distance parameters $\Delta$ and $\Delta_\text{PDE}$ which determine how quickly the algorithm expands the solution interval. When $\sigma^2$ is small then the interval is expanded more slowly and the algorithm may require more iterations to converge. Too large values of $\sigma^2$ can result in adding new regions in the training procedure too early. The distance hyperparameters $\Delta$ and $\Delta_\text{PDE}$ have a similar effect. However, for the considered systems, the model performance is stable with little sensitivity to the values of these hyperparameters as well as to the number $\Kpde$ of collocation points and the number of networks in the ensemble. 

\begin{table*}[ht]
\caption{
The results obtained with the proposed 'PINN Ensemble' algorithm using different hyperparameter values. We report the same metric for the same systems as in Table~\ref{tab:results}.
The superscript $^{(n)}$ indicates how many runs among the 10 runs did not converge after a fixed number of updates. We assume that the algorithm has converged if more than 95\% of the sampled points have been added to the set used to compute loss $\Lpde$.
}
\label{tab:ens_hyperparams}
\footnotesize
\begin{center}
\resizebox{\textwidth}{!}{
\begin{tabular}{@{}c|cc|ccc|ccc}
\multirow{2}{*}{} & \multicolumn{2}{c|}{Convection, $\beta$} & \multicolumn{3}{c|}{Reaction, $\rho$} & \multicolumn{3}{c}{Reaction-diffusion, $\nu$}
\\
& $\beta=30$ & $\beta=40$ & $\rho=5$ & $\rho=6$ & $\rho=7$ & $\nu=2$ & $\nu=3$ & $\nu=4$
\\
\hline
default
          & 7.00 (2.7) & 12.6 (3)
          & 7.21 (1.6) & 11.0 (2) & 13.4 (4)
          & 6.93 (.3) & 6.68 (.4) & 6.63 (.3)
\\
\hline
\multicolumn{9}{c}{}\\
\multicolumn{9}{c}{Varying the threshold for the ensemble disagreement, default $\sigma^2=0.0004$}
\\
\hline
$\sigma^2=10^{-4}$
          & 8.15 (2.7)$^{(1)}$ & 14.0 (2)
          & 7.16 (1.6) & 11.3 (2) & 12.3 (3)$^{(3)}$
          & 6.67 (.4) & 6.73 (.6) & 6.60 (.5)
\\
$\sigma^2=10^{-3}$
          & 8.11 (2.0) & 14.5 (4)
          & 7.16 (1.5) & 11.3 (2) & 13.0 (4)
          & 6.68 (.3) & 6.79 (.4) & 6.63 (.4)
\\
\hline
\multicolumn{9}{c}{}\\
\multicolumn{9}{c}{Varying the distances which determinate candidate collocation points,
default $\Delta_\text{PDE} = 0.1$, $\Delta = 0.05$}
\\
\hline
$\Delta_\text{PDE} = .07$
          & 10.1 (4.4) & 17.6 (10)$^{(2)}$
          & 6.80 (1.1) & 11.5 (3) & 10.7 (3)$^{(2)}$
          & 7.03 (.3) & 6.76 (.6) & 6.68 (.6)
\\
$\Delta_\text{PDE} = .2$
          & 5.33 (0.9) & 11.5 (2)
          & 7.61 (2.) & 12.7 (3) & 11.6 (4)
          & 7.13 (.4) & 6.89 (.4) & 6.49 (.4)
\\
$\Delta = .07$
          & 7.82 (1.6) & 13.2 (2)
          & 7.67 (1.5) & 12.0 (2) & 15.3 (4)
          & 6.70 (.6) & 6.64 (.5) & 6.63 (.4)
\\
\doublecell{$\Delta_\text{PDE} = .25$ \\ \& $\Delta = .125$}
          & 6.25 (1.6) & 13.0 (2)
          & 8.42 (2.) & 14.2 (3) & 14.5 (8)
          & 6.97 (.4) & 6.88 (.6) & 6.75 (.4)
\\
$\Delta = 10^5$
          & 8.98 (1.9) & 13.7 (2)
          & 8.33 (1.6) & 13.6 (2) & 18.4 (3)
          & 6.69 (.2) & 6.91 (.6) & 6.62 (.4)
\\
$\Delta_\text{PDE} = 10^5$
          & \textit{11.1 (1.2)} & \textit{16.9 (3)}
          & \textit{8.22 (1.5)} & \textit{56 (71)} & \textit{36.3 (53)}
          & \textit{455 (143)} & \textit{242 (164)} & \textit{74 (82)}

\\
\hline
\multicolumn{9}{c}{}\\
\multicolumn{9}{c}{Varying the prediction error which determinate candidate collocation points,
default $\epsilon = 0.001$}
\\
\hline
$\epsilon=5 \cdot 10^{-4}$
          & 8.94 (2.3) & 13.9 (6)
          & 7.13 (1.5) & 11.2 (2) & 12.9 (4)$^{(1)}$
          & 6.66 (.3) & 6.93 (.7) & 6.61 (.4)
\\
$\epsilon= 10^{-2}$
          & 7.65 (1.7) & 12.3 (2)
          & 7.15 (1.5) & 11.4 (2) & 13.0 (4)$^{(1)}$
          & 6.72 (.4) & 6.93 (.7) & 6.63 (.3)
\\
$\epsilon= 10^{5}$
          & 7.19 (1.4) & 12.2 (2)
          & 7.11 (1.5) & 11.4 (2) & 12.9 (4)$^{(1)}$
          & 6.72 (.4) & 6.93 (.7) & 6.61 (.4)
\\
\hline
\multicolumn{9}{c}{}\\
\multicolumn{9}{c}{Varying the number of the collocation points, default $\Kpde=1000$}
\\
\hline
$\Kpde=5000$
          & 8.02 (1.9) & 14.1 (2)
          & 6.75 (1.4) & 10.7 (2) & 10.4 (3)
          & 6.88 (.4) & 6.52 (.5) & 6.55 (.5)
\\
$\Kpde=10^4$
          & 7.71 (1.7) & 16.8 (4)
          & 7.00 (1.3) & 9.89 (2) & 10.3 (4)
          & 6.94 (.4) & 6.71 (.5) & 6.55 (.4)
\\
\hline
\multicolumn{9}{c}{}\\
\multicolumn{9}{c}{Number of training epochs for the first iteration, default $N_1=5000$}
\\
\hline
$N_1=10^3$
          & 7.12 (1.0) & 15.6 (3)
          & 7.86 (1.3) & 12.6 (2) & 14.8 (2)$^{(2)}$
          & 6.89 (.3) & 6.45 (.4) & 6.58 (.2)
\\
$N_1=10^4$
          & 10.8 (7) & 15.0 (8)$^{(1)}$
          & 6.26 (1.4) & 9.18 (2) & 9.73 (3.2)
          & 6.79 (.3) & 6.67 (.4) & 6.66 (.5)
\\
\hline
\multicolumn{9}{c}{}\\
\multicolumn{9}{c}{Number of training epochs between adding new points, default $N=1000$}
\\
\hline
$N=700$
          & 7.64 (1.4) & 15.1 (7)          
          & 7.53 (1.7) & 12.1 (2) & 14.3 (3)$^{(1)}$          
          & 6.91 (.4) & 6.66 (.4) & 6.65 (.2)
\\
$N=2000$
          & 9.95 (1.7) & 14.4 (5)
          & 6.62 (1.3) & 9.99 (2) & 11.8 (3)
          & 6.96 (.4) & 6.87 (.4) & 6.54 (.5)
\\
\hline
\multicolumn{9}{c}{}\\
\multicolumn{9}{c}{Number of networks, default $L=5$}
\\
\hline
$L=3$
          & 8.96 (3.7) & 12.2 (4)
          & 6.45 (1.4) & 11.6 (4) & 14.4 (6)$^{(3)}$
          & 6.83 (.3) & 6.60 (.3) & 6.75 (.6)
\\
$L=7$
          & 6.68 (1.6) & 12.3 (2)
          & 6.07 (.81) & 11.4 (2) & 11.9 (3)$^{(5)}$
          & 6.88 (.2) & 6.78 (.4) & 6.57 (.2)
\\
$L=10$
          & 7.94 (1.4) & 13.6 (2)
          & 6.48 (.58) & 10.6 (.9) & 12.5 (3)
          & 7.05 (.3) & 6.72 (.1) & 6.59 (.2)
\\
\hline

\end{tabular}
}
\end{center}
\end{table*}

\subsection{More experimental results}
In this section we report additional experimental results for reaction (Eqs.~\ref{eq:react}--\ref{eq:ibreact}) with $\rho=6$ and reaction-diffusion (Eqs.~\ref{eq:rd}--\ref{eq:ibrd}) with $\nu=2$.

\begin{table*}[hb]
\caption{
The means (and standard deviations) across 10 random seeds of the relative $l_2$ errors in Eq.~\ref{eq:r} for reaction (Eqs.~\ref{eq:react}--\ref{eq:ibreact}) and reaction-diffusion (Eqs.~\ref{eq:rd}--\ref{eq:ibrd}) systems. All numbers should be multiplied by $10^{-3}$.
The rows marked with $^*$ show the results obtained with $\Kpde=25600$ collocation points otherwise $\Kpde=1000$.
In \textBF{bold}: the best three results.
}
\label{tab:results_extra}
\begin{center}
\resizebox{0.6\textwidth}{!}{ 
\begin{tabular}{@{}c|c|c}
\multirow{2}{*}{Method} & \multicolumn{1}{c|}{Reaction, $\rho$} & \multicolumn{1}{c}{Reaction-diffusion, $\nu$}
\\
& $\rho=6$ & $\nu=2$ 
\\
\hline
PINN, LBFGS %
          & 993 (2)
          & 608 (469)
\\
PINN, Adam
          & 592 (459)
          & 7.01 (0.74) 
\\
PINN, Adam+LBFGS
          & 678 (433)
          & \textBF{6.83 (0.03)} 
\\
SA-PINN %
        & 8.12 (1.40)
        & 6.97 (0.54) 
\\
Causality
          & 102 (35)
          & 678 (5) 
\\
Causality$^*$ %
          & 18.9 (4.1)
          & 7.06 (0.27) 
\\
bc-PINN, Adam
        & \textBF{5.66 (2.02)}
        & 7.11 (1.35) 
\\
bc-PINN, Adam+LBFGS
        & \textBF{2.96 (1.25)}
        & 6.89 (0.08) 
\\
\hline
\hline
PL, Adam (our)
          & 38.5 (3.5)
          & 8.83 (0.42) 
\\
PL, Adam+LBFGS  (our)
          & \textBF{4.17 (0.58)}
          & \textBF{6.83 (0.02)} 
\\
Ens, Adam (our) %
          & 11.0 (2.3)
          & 6.93 (0.32) 
          
\\
Ens, Adam+LBFGS (our) %
          & 9.73 (2.02) & \textBF{6.85 (0.02)}
\\
\end{tabular}
}
\end{center}
\end{table*}

\end{document}